\documentclass{article}
\usepackage{arxiv}
\usepackage{hyperref}
\usepackage[utf8]{inputenc}
\usepackage{easyReview}
\usepackage{url}

\usepackage[utf8]{inputenc}

\usepackage{verbatim}
\usepackage{ifthen}
\usepackage{xspace}
\usepackage{graphicx}
\usepackage{multimedia} 
\usepackage{pgf}
\usepackage{rotating}

\usepackage{amsmath,amssymb}

\usepackage[vlined,ruled]{algorithm2e}
\usepackage{tikz}

\usepackage{color}
\usepackage{xcolor}
\usepackage{tcolorbox}
\usepackage{colortbl}
\usepackage[makeroom]{cancel}

\usepackage{ucs}
\usepackage{comment}
\usepackage{fancybox}
\usepackage{caption}
\usepackage{subcaption}
\usepackage{bm}
\usepackage{rotating}

\usepackage{array, makecell}
\usepackage{multirow}
\usepackage{colortbl}
\usepackage{tcolorbox}
\usepackage{xspace}

\newcommand{\burl}[1]{\structure{\url{#1}}}
\newcommand{\mymath}[1]{\ensuremath{#1}\xspace}
\newcommand{\mymathbf}[1]{\mymath{\mathbf{#1}}}

\def\bbbr{{\rm I\!R}} 
\newcommand{\R}{{\bbbr}{}}

\newcommand\no[1]{}
\newcommand\wi[1]{$\circ$}
\newcommand\bu[1]{$\bullet$}
\newcommand\ot[1]{$\star$}
\newcommand\bo[1]{$\bullet\star$}

\newcommand{\A}{\mymathbf{A}}

\newcommand{\G}{\mymathbf{G}}
\newcommand{\Om}{\mymathbf{\Omega}}
\newcommand{\g}{\mymathbf{g}}

\newcommand{\Ob}{\mymathbf{O}}

\newcommand{\Rew}{\mymathbf{R}}
\newcommand{\T}{\mymathbf{T}}

\newcommand{\poet}{{\sc poet}\xspace}

\newcommand{\rien}[1]{}

\DeclareGraphicsExtensions{.jpg,.mps,.pdf,.png,.gif}

\newcommand{\mycite}[1]{
\par\medskip
{\em ``#1"}
\par\medskip
}

\definecolor{myred}{rgb}{0.8,0,0}
\definecolor{mygreen}{rgb}{0,0.6,0}
\definecolor{myblue}{rgb}{0,0,0.7}

\definecolor{DarkGray}{gray}{0.9}
\definecolor{MediumGray}{gray}{0.75}
\definecolor{LightGray}{gray}{0.5}

\newcounter{ques} 
\newcommand{\ques}{\arabic{ques}}

\newcounter{ass} 
\newcommand{\ass}{\arabic{ass}}

\newcounter{cpb} 
\newcommand{\cpb}{\arabic{cpb}}
\newenvironment{defpb}[1]{\refstepcounter{cpb}\begin{tcolorbox}[colback=blue!10!white]{\bf Definition Pro\cpb:} #1}{\end{tcolorbox}}

\newcounter{ck} 
\newcommand{\ck}{\arabic{ck}}
\newenvironment{keydef}[1]{\refstepcounter{ck}\begin{tcolorbox}[colback=blue!30!white]{\bf Key definition \ck:} #1}{\end{tcolorbox}}

\newcounter{csol} 
\newcommand{\csol}{\arabic{csol}}
\newenvironment{soldef}[1]{\refstepcounter{csol}\begin{tcolorbox}[colback=red!30!white]{\bf Definition Sol\csol:} #1}{\end{tcolorbox}}

\usepackage{makecell}

\title{A Definition of Open-Ended Learning \\ for Goal-Conditioned Agents}

\author{
\href{https://orcid.org/0000-0002-8544-0229}%
{\includegraphics[scale=0.06]{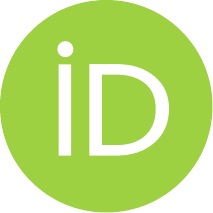}%
\hspace{1mm}Olivier Sigaud$^{(1)*}$}, %
\href{https://orcid.org/0000-0002-1277-4447}%
{\includegraphics[scale=0.06]{orcid.pdf}%
Gianluca Baldassarre$^{(3)*}$}, %
\href{https://orcid.org/0000-0003-0212-427X}%
{\includegraphics[scale=0.06]{orcid.pdf}%
\hspace{1mm}C\'{e}dric Colas$^{(2, 5)}$}, %
\\
\href{https://orcid.org/0000-0002-2920-4539}%
{\includegraphics[scale=0.06]{orcid.pdf}%
\hspace{1mm}{\bf St\'{e}phane Doncieux$^{(1)}$}}, %
\href{https://orcid.org/0000-0002-6807-524X}%
{\includegraphics[scale=0.06]{orcid.pdf}%
{\bf Richard Duro$^{(4)}$}}, %
\href{https://orcid.org/0000-0002-1277-130X}%
{\includegraphics[scale=0.06]{orcid.pdf}\hspace{1mm}%
{\bf Pierre-Yves Oudeyer$^{(2)}$}}, %
\\
\href{https://orcid.org/0000-0001-8626-1938}%
{\includegraphics[scale=0.06]{orcid.pdf}\hspace{1mm}%
{\bf Nicolas Perrin-Gilbert$^{(1)}$}},%
\href{https://orcid.org/0000-0002-1277-4447}%
{\includegraphics[scale=0.06]{orcid.pdf}\hspace{1mm}%
{\bf Vieri Giuliano Santucci$^{(3)}$}}, %
\\
  (1) Sorbonne Universit\'e, CNRS, Institut des Syst\`emes Intelligents et de Robotique,\\
  F-75005 Paris, France\\
  (2) INRIA Bordeaux Sud-Ouest, \'equipe FLOWERS, France\\
  (3) Institute of Cognitive Sciences and Technologies, CNR, Rome, Italy\\
  (4) CITIC, Universidade da Coru\~{n}a, Spain\\
  (5) Massachusetts Institute of Technology, Cambridge, USA\\
  (*) Equally contributing authors, the other authors are listed in alphabetical order\\
Correspondence to Olivier.Sigaud@isir.upmc.fr}

\begin{document}

\maketitle

\begin{abstract}
A lot of recent machine learning research papers have ``open-ended learning'' in their title. But very few of them attempt to define what they mean when using the term. Even worse, when looking more closely there seems to be no consensus on what distinguishes open-ended learning from related concepts such as continual learning, lifelong learning or autotelic learning. In this paper, we contribute to fixing this situation. After illustrating the genealogy of the concept and more recent perspectives about what it truly means, we outline that open-ended learning is generally conceived as a composite notion encompassing a set of diverse properties.
In contrast with previous approaches, we propose to isolate a key elementary property of open-ended processes, which is to produce elements from time to time (e.g., observations, options, reward functions, and goals), over an infinite horizon, that are considered novel from an observer's perspective. From there, we build the notion of open-ended learning problems and focus in particular on the subset of open-ended goal-conditioned reinforcement learning problems in which agents can learn a growing repertoire of goal-driven skills. Finally, we highlight the work that remains to be performed to fill the gap between our elementary definition and the more involved notions of open-ended learning that developmental AI researchers may have in mind.
\end{abstract}

\section{Introduction}

Most existing software agents and robots suffer from insufficient versatility, limiting the potential introduction of these agents and robots in our everyday life \cite{plappert2018multi}. In most cases, some expertise is required from a human engineer to design their behavior in anticipation of the situations they may encounter. As a consequence, these agents cannot address novel or unforeseen situations. 

An alternative to this specific expertise requirement would be to build an agent which would be ready to address {\em any} problem, fulfilling the long-standing dream of Artificial General Intelligence (AGI) \cite{fjelland2020general}.The corresponding design effort would be extraordinary, so there is a consensus that the way to reach AGI should be through learning. But the AGI perspective does not account for one of the core limitations of human intelligence: some humans can solve problems that others cannot, and vice versa, because humans have a limited lifespan for learning whereas being proficient in all potential tasks would require a potentially infinite amount of learning time.

Taking into account the above core limitation, open-ended learning (OEL) is a framework where autonomous agents face new tasks and learn how to solve them over their lifespan, driven by some intrinsic motivations or external guidance. Doing so, they build their own competence along some {\em developmental trajectory} which endows them with some capabilities, but not all capabilities that one may dream of \cite{Lungarella2003DevelopmentalRobotics}.
Their developmental trajectory consists of a curriculum of tasks, and the curriculum learning challenge consists in finding when to learn which tasks so as to better extend the agent's capabilities.

Though the OEL topic has been around for more than a decade (e.g., see the Intrinsically Motivated Open-ended Learning Workshops and Community\footnote{IMOL - Intrinsically Motivated Open-ended Learning Workshops and Community: \url{https://www.imol-community.org/archive/}} and Section~\ref{sec:lit}), the interest for this framework is growing very fast in the machine learning community, as illustrated by more than 200 machine learning papers using the term in 2022, the emergence of new workshops (e.g. ALOE\footnote{Agent Learning in Open-Endedness (ICLR 2022 Workshop): \url{https://sites.google.com/view/aloe2022}}) or even the very related COLLAS top-tier conference\footnote{Conference on Lifelong Learning Agents: \url{https://lifelong-ml.cc/}}. This growth is simultaneous with the emergence of related topics in Computational Neuroscience (e.g., see \cite{niv2019learning,rmus2021role}).

In this paper we start by showing that, despite a growing number of research works using the term ``open-ended learning'', there have been very few attempts at providing a formal definition for the corresponding concept. This results in a lack of consensus on what OEL truly is and in a risk of confusion with several very related topics such as lifelong learning, continual learning or autotelic learning.

\section{Defining OEL: Elements from the literature}
\label{sec:lit}

In this section, we draw from the literature some elements of a definition of OEL. We investigate the genealogy of the notion through its first occurrences in Artificial Intelligence (AI) and Artificial Life (AL) papers and we highlight the conceptual shift that arose in more recent papers. This investigation helps us set the stage for providing a better definition than the few existing ones in a subsequent section.

\subsection{Genealogy of the notion}

Searching for the very first instances of AI or AL works mentioning open-endedness is difficult because OEL is also a concept used in education sciences, with a far longer history. 
The first reference we could find where the term ``open-ended'' referred to intelligent machines within a cognitive science context is from \cite{Reader1969}:
\mycite{So far as is known, the range of tasks which the human intellect can master is open-ended (infinite), and therefore an intelligent machine can never be proved to be intelligent by comparing its task performance with that of the human intellect since the process would not terminate. [...] The only way a machine can continue to impersonate the open-ended ability of the human intellect is by also being open-ended. [...] The human intellect has an open-ended ability because it is capable of learning and the intelligent machine must also be capable of learning.}

This view for which an intelligent machine should learn forever as humans do is clearly reminiscent of Turing's claim in his seminal paper that the process of making a machine intelligent should follow ``the normal teaching of a child'' \cite{turing1950computing}.

Much later, an interesting early appearance of open-endedness in a pure AI paper is \cite{meyer1991artificial}, a text from the founders of the ``From animals to animats'' community at the time where the corresponding conference emerged.
They mention ``an open-ended space of network architectures'' in a paper dedicated to the artificial evolution of behaviours. The notion of ``open-ended evolution'' soon became the topic of several papers in the AL community, where the goal was to explain how evolution can generate increasingly complex and capable creatures in a context where most evolutionary algorithms were disappointingly converging to some unsatisfactory fixed-point or cyclic patterns. For instance, \cite{standish2003open} mentions such discussions as central to the AL conferences in the early 2000's, and open-ended evolution is still an active topic in the AL community, see for example \cite{taylor2016open, packard2019overview}.
To extract elements of a definition from this line of research, we retain the definition from \cite{standish2003open}, where we find:
\mycite{The issue of open-ended evolution can be summed up by asking under what conditions will an evolutionary system continue to produce novel forms.}

Importantly, we see that the focus is on the conditions on the environment rather than on the evolution process itself (e.g., see also \cite{soros2014identifying}, for similar conclusions) and that the expected outcome is the production of novelty.

One perspective on the conditions for open-ended evolution or OEL is thus to consider an  rich enough environment which provides an open-ended sequence of problems. A similar idea of generating sequences of problems dates back to \cite{schmidhuber2013powerplay} where the author defines ``Open-Ended PowerPlay''. The goal of the approach is the ``Invention of new problems'' \cite{srivastava2012continually} to continually challenge a learning agent. This is much more closely related to what the machine learning community now means with OEL, but note that the focus is on generating problems in the environment rather than on the necessary mechanisms to solve them. This work is often cited in the machine learning works interested in the automated generation of curricula, and particularly in those which focus on generating a sequence of challenging problems in some environment rather than on solving them. For instance, this is the case of the \poet algorithm \cite{wang2019poet, wang2020enhanced} where the open-ended invention of challenging environments is conceptualized as separated from the agent:
%
%
\mycite{How can progress in machine learning and reinforcement learning be automated to generate its own never-ending curriculum of challenges without human intervention? The recent emergence of quality diversity (QD) algorithms offers a glimpse of the potential for such continual open-ended invention. 
[...] The Paired Open-Ended Trailblazer (POET) algorithm introduced in this paper combines these principles to produce a practical approach to generating an endless progression of diverse and increasingly challenging environments while at the same time explicitly optimizing their solutions.}

This perspective about open-ended invention is now widely adopted in the machine learning community, and can be seen as one of the most prevalent understandings of what OEL truly means, for example see
\cite{dharna2020co, bontrager2021learning, kepes2022chemgrid, dharna2022watts}.
This view has been recently strengthened and further broadened with the proposal of a theoretical perspective claiming that the current challenges of machine learning, including the production of general artificial intelligence, can be addressed shifting focus from learning algorithms to \textit{generalized exploration}.
This allows both reinforcement learning (RL) and supervised learning (SL) algorithms to automatically generated new data \cite{jiang2023general}: 

\mycite{Importantly, generalized exploration is a necessary objective for maintaining open-ended learning processes, which in continually learning to discover and solve new problems, provides a promising path to more general intelligence.}

Within this framework, the authors propose that promising approaches involve processes for the open-ended generation of novel data that maximize the agent's learning potential while also leading it to explore areas of the problem space that are at the same time novel and grounded (i.e., relevant for the use of the system).
This approach might for example involve the open-ended automatic generation of novel and relevant tasks in RL, for example based on parameterised environments, and novel and relevant new data in SL, for example based on the generation of synthetic training data.

A different perspective, pioneered by \cite{Weng2000Computational, weng2001autonomous}, shifts the focus of the ``openness'' from the environment to the agent:

\mycite{What is autonomous mental development? With time, a brain-like natural or an artificial embodied system, under the control of its intrinsic developmental program (coded in the genes or artificially designed) develops mental capabilities through autonomous real-time interactions with its environments (including its own internal environment and components) by using its own sensors and effectors. Traditionally, a machine is not autonomous when it develops its skills, but a human is autonomous throughout its lifelong mental development. [...] A mental developmental process is also an open-ended cumulative process} 

In Weng's perspective, the autonomous progressive learning of the agent should be guided by an internal \textit{developmental program}. This program is characterised by various elements that are relevant for OEL:
``\textit{Sensor-specific and effector-specific; task-nonspecific; tasks unknown at programming time; generate representation automatically; animal-like online learning; open-ended learning of more new tasks.}''
This approach led to the birth of the new field of ``Developmental Robotics'' and the related ``Development and Learning'' interdisciplinary community and conference \cite{Lungarella2003DevelopmentalRobotics}.

In this context, other seminal works identified the key computational ingredients that allow the actual implementation of the developmental program, namely \textit{intrinsic motivations} \cite{Barto2004IntrinsicallyMotivatedLearning,Chentanez2004IntrinsicallyMotivated}.
This is also the case of \cite{prince2005ongoing} who introduces a notion of {\em ongoing emergence} which is explicitly related to open-ended learning and intrinsic motivations.
These works managed to ignite a research effort directed to systematically draw concepts on different intrinsic motivations from psychology, and translate them into specific machine-learning algorithms \cite{Barto2004IntrinsicallyMotivatedLearning}: 
\mycite{Psychologists distinguish between extrinsic motivation, which means being moved to do something because of some specific rewarding outcome, and intrinsic motivation, which refers to being moved to do something because it is inherently enjoyable. Intrinsic motivation leads organisms to engage in exploration, play, and other behavior driven by curiosity in the absence of explicit reward. [...] Although these arguments are compelling, developmental approaches to artificial agent design have been slow to penetrate the mainstream of the machine learning community.}

Intrinsic motivations can thus support the possibly-open \textit{autonomous} acquisition of knowledge and skills:

\mycite{According to this approach, an agent undergoes an extended developmental period during which collections of reusable skills are autonomously learned that will be useful for a wide range of later challenges.}

These works followed previous pioneering works on specific intrinsic motivation mechanisms which initially were overlooked \cite{Schmidhuber1990Making,
Schmidhuber1991Curious}. 
The renewed research effort led to connect educational science and developmental psychology concepts of OEL in children to the notion of intrinsic motivations in machine learning and robots \cite{kaplan2007search}.
In addition, it led to distinguish and formalise different classes of intrinsic motivations, in particular related to \textit{prediction}, \textit{novelty}, and \textit{competence} \cite{Oudeyer2007WhatisIntrinsic,Barto2013Noveltyorsurprise}.
Intrinsic motivations can thus form the `motivational engine' guiding \textit{autonomous} open-ended learning  in organisms and robots \cite{Baldassarre2013IntrinsicBook}.

Another critical step for the definition of OEL agents was the investigation of the relation between intrinsic motivations and evolution, which we have seen to be two key areas where open-endedness can manifest.
In general, evolution is proposed to lead to the emergence of both extrinsic motivations (those serving typical biological needs such as hunger and sex) and intrinsic motivations as both can guide the acquisition of behaviour adapted to the environmental conditions
\cite{Schembri2007EvolutionIntrinsically,
Singh2010IntrinsicallyEvolutionary}.
In organisms, both types of motivations are supported by brain mechanisms responding to different general principles \cite{Baldassarre2011IntrinsicBiological}. 
In particular, \textit{extrinsic motivations} support the acquisition of material resources having a direct adaptive advantage, and to this purpose monitor visceral body states. Instead, intrinsic motivations, emerged later in evolution, support the acquisition of knowledge and skills which are only later useful to increase fitness, and thus monitor the brain \textit{information gain}. For this reason, only intrinsic motivations, if suitably translated into algorithms, have the ``\textit{potential to produce open-ended learning machines and robots}'' \cite{Baldassarre2011IntrinsicBiological}. This idea led to the start of the IMOL Workshop series mentioned above.

Aside from intrinsic motivations, research has highlighted that a second key element can foster OEL, namely \textit{goals}. Goals are internal representations of desired states of the environment that can guide the agent's action. Goals represent a key element in classic symbolic AI systems, for example to support planning \cite{russell2016artificial}, but have been less employed in ML systems.
A main exception is the RL option framework, where goals might be associated to the termination condition of options \cite{Sutton999AbstractionRL}, or \textit{goal-conditioned policies}, where RL action policies are goal indexed \cite{Baldassarre2001Aplanningmodular,liu2022goal, colas2022autotelic}.
Within this context, the potential relevance of goals for OEL was highlighted in \cite{Barto2004IntrinsicallyMotivatedLearning}, where the learning of the agent was guided by the autonomous generation of options (hence goals) when a ``salient event'' was encountered.
Later, self-generated goals have been indicated to be a fundamental means usable by OEL agents to autonomously generate curricula driven by competence-based intrinsic motivations
\cite{Santucci2012competence,Santucci2016GRAIL}. The EU-funded project GOAL-Robots\footnote{GOAL-Robots - Goal-based Open-ended leaning Autonomous Robots: https://www.goal-robots.eu/} pivoted on this idea to build \textit{GOAL Agents}, where `GOAL' stands for `Goal-based Open-ended Autonomous Learning'. The capacity of OEL agents to autonomously generate, discover or select goals has been recently called \textit{autotelic} \cite{colas2022autotelic}, and plays an important role in the framework proposed here.

\begin{figure*}[!ht]
    \centering
    \includegraphics[width=0.8
    \textwidth]{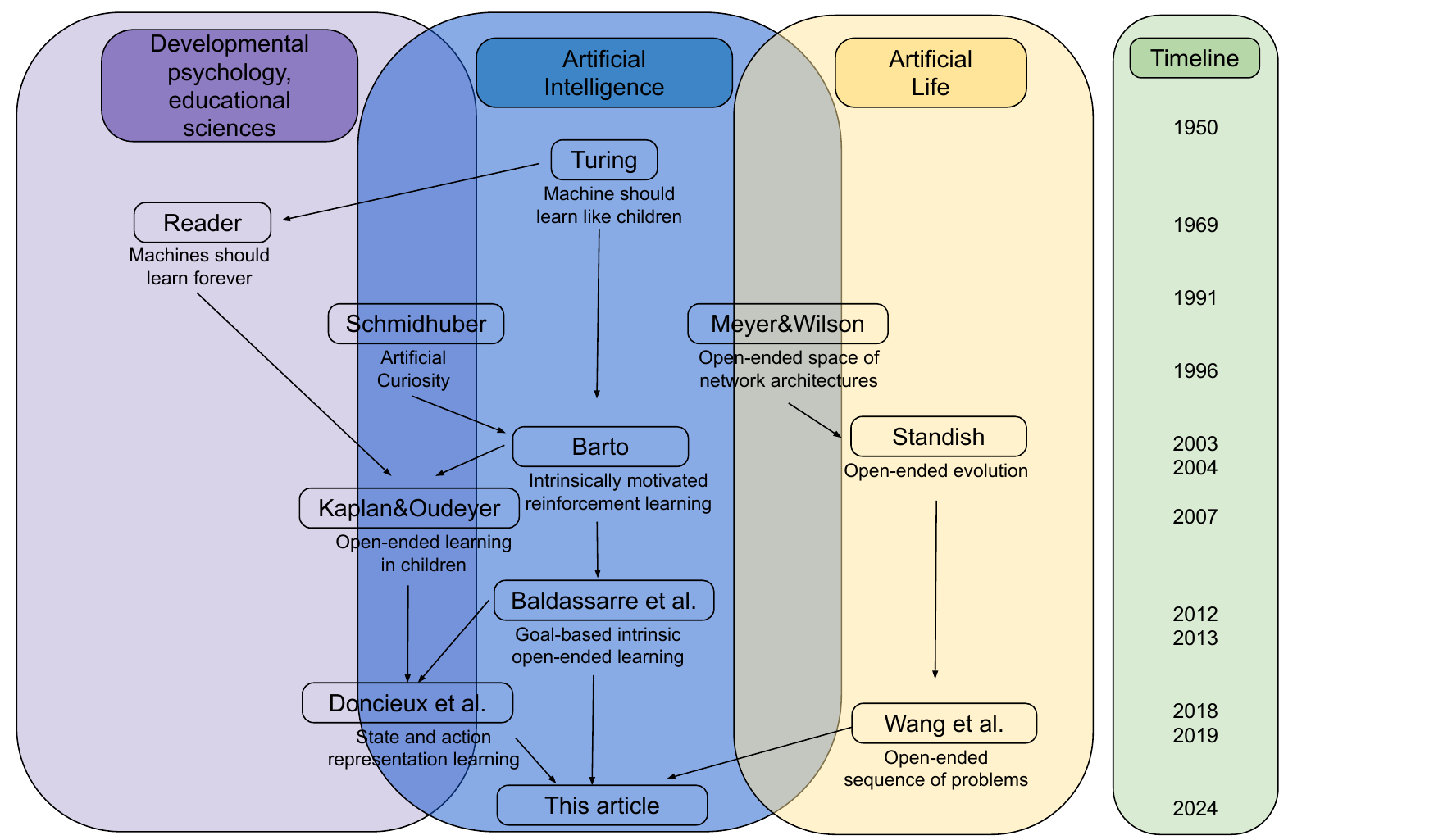}
    \caption{Schematic overview of the genealogy of the notion of open-endedness in the Developmental AI and Artificial Life literature.}
    \label{fig:genealogy} 
\end{figure*}

We summarize the above overview of the genealogy of the notion of open-endedness in the AI/AL literature in \figurename~\ref{fig:genealogy}.
From the historical investigation above, we can retain that the open-endedness property can be bound to the environment or to the agent.
In the latter case, open-ended learning relies on intrinsic motivations able to drive the autonomous unbounded learning of increasingly complex and new knowledge and skills.
In this case, goals have emerged as a key means to support open-ended learning.

\subsection{Current perspectives}

Though all the perspectives considered in the previous section still correspond to active research lines, we observe a renewal of the questions related to open-endedness in the more recent literature. For instance, though they do not define what they mean by OEL, \cite{fan2022minedojo} have OEL environments, tasks, goals, and task suites, the main idea being to have a ``wide variety'' of such elements.

A relevant recent paper from the ``Open-ended learning group'' at DeepMind seems to adopt the above perspective. 
Though the term ``open-ended'' hardly appears apart from
the title, introduction, and conclusion, we can find in \cite{stooke2021open} the following claim:

\mycite{We show that through constructing an open-ended learning process, which dynamically changes the training task distributions and training objectives such that the agent never stops learning, we achieve consistent learning of new behaviours.}

One can see that an OEL process should provide a goal distribution shift, 
or curriculum, from which the agent can continually learn to achieve new goals. As outlined in Section~\ref{sec:combi}, this may be confused with continual or lifelong learning. The same is true of a more recent article from the same company \cite{team2023human}.

The work \cite{doncieux2018open} present another perspective that is relevant for our attempt to build a stronger definition of OEL. The authors define an {\em OEL process} as a process building appropriate MDPs depending on the current situation and the external reward function. This perspective
has two major features. First, it is the only paper we found that attempts to provide an explicit definition of OEL. Second, it departs from all other existing frameworks by considering that each task should come with its own state and action spaces. This is in sharp contrast with all the multitask RL frameworks we know about, where a unique policy or set of policies with the same input-output format are used, implying that the state and action spaces are the same for all tasks. As outlined in a previous work tackling the issue of representations within the RL framework \cite{konidaris2019necessity}, having a necessary and sufficient representation specific to each task corresponds to the necessity of state and action {\em abstraction}: it is easier to solve a task if we abstract away all sensory information and potential actions that are not relevant for task achievement.

The same idea can be found in a recent trend in computational neuroscience research which focuses on the executive functions that help us determine an adequate representation of the task we want to solve \cite{niv2019learning, rmus2021role}.
In \cite{doncieux2018open}, the capability to abstract away adequate state and action spaces for each task is based on a more general representational redescription capability inspired by \cite{karmiloff1994beyond}.

However, the definition of OEL provided in \cite{doncieux2018open} is not entirely satisfactory. A key point is that the framework considers that, when an agent addresses an open-ended sequence of tasks, each task comes with its own externally defined reward function. But if the tasks cannot be anticipated, the corresponding reward functions cannot be provided in advance by the agent designers.
Moreover, and most importantly, although it is an interesting condition, it is in no way obvious why abstraction and representational redescription capabilities would be a necessary condition for exhibiting OEL capabilities.
Finally, in \cite{doncieux2018open} goals and motivations are built from the reward. Instead, intrinsic motivations frameworks do the contrary by deriving goals from intrinsic motivations and rewards from intrinsic motivations and goal achievement \cite{Barto2004IntrinsicallyMotivatedLearning, Santucci2016GRAIL, colas2022autotelic}.

This final remark can be generalized to many works cited above. Indeed, a key issue with previous attempts at defining OEL is that they all try to bind the concept to a large conjunction of properties, such as being autonomous, being endowed with intrinsic motivations or being capable of state abstraction. Relying on such conjunctions does not help pinpointing what is specific to OEL, which we believe is a key requirement for faster progress in designing agents capable of this property.

The work that best responds to this remark is \cite{romero2023perspective}, where the authors put forward the Lifelong Open-Ended Learning Autonomy (LOLA) framework. We think that by isolating OEL from lifelong learning and autonomous learning, the LOLA framework of \cite{romero2023perspective} takes a step in the right direction. However, for the OEL component of their framework they refer to \cite{doncieux2018open} without providing a proper definition, though the referred paper does not provide a satisfactory OEL definition. 

\begin{figure*}[!ht]
    \centering
    \includegraphics[width=0.95\textwidth]{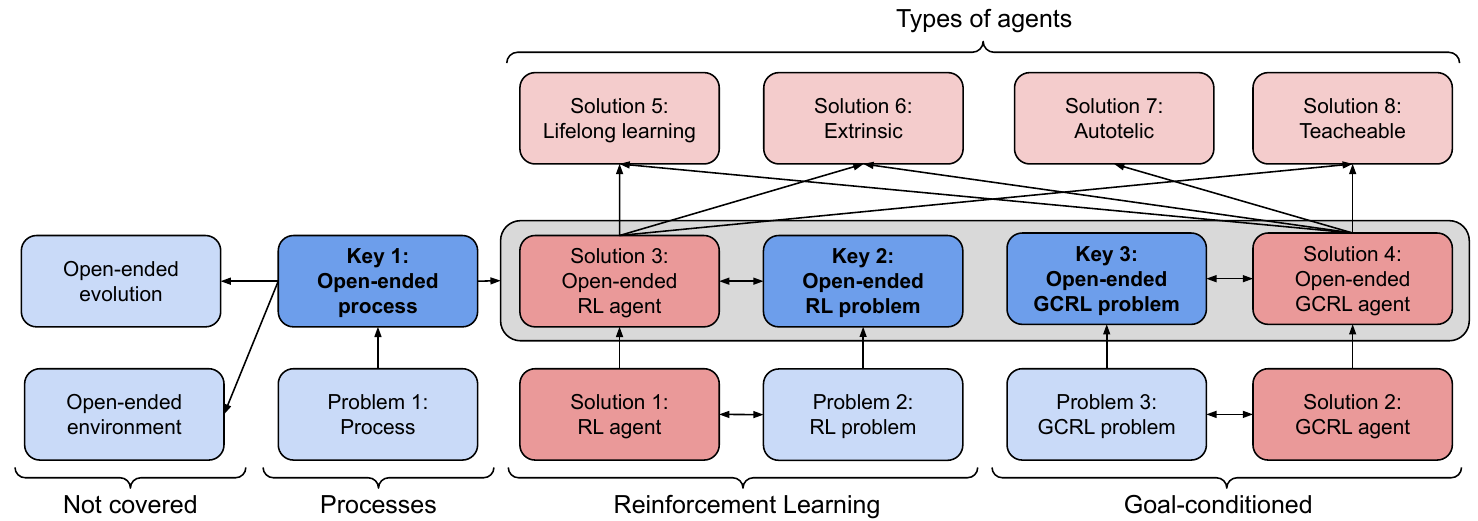}
    \caption{The set of definitions used in this paper. In blue, notions related to defining the open-ended goal-conditioned RL \textit{problems} (key definitions are darker). In red, notions related to defining open-ended GCRL agents, that is \textit{solutions}. An arrow expresses that a definition depends on another.}
    \label{fig:framework} 
\end{figure*}

Thus the key contribution in Section~\ref{sec:def} consists in providing such a core definition, on which we hope to build a solid OEL framework. Then in Section~\ref{sec:combi}, we combine the just defined OEL property with other properties mentioned in the literature. In particular, we show how the property can be combined with lifelong learning and that it removes the need for the continual RL framework of \cite{abel2023definition}. In addition, we discuss the fact that an OEL agent might be autotelic or it could be teachable (that is, it receives goals from the environment). Finally, in Section~\ref{sec:limits} we discuss the limits of our proposal. We summarise all these elements and their relations in \figurename~\ref{fig:framework}, which expresses the relationships between the definitions we propose.

\section{Open-ended learning: a definition}
\label{sec:def}

Our investigation of the literature about the definition of OEL in AI and ALife has revealed that some authors were trying to define {\em OEL problems} whereas others were focusing on defining {\em OEL solutions}. The coexistence of both perspectives is not surprising as it is unclear whether the conditions for the emergence of OEL are more on the side of the environment --~which defines the problem the agent should solve~-- or on the side of the agent --~where a class of agents defines the solution to the OEL problem. In this paper, we want to define OEL problems in the most general way, so we consider both perspectives. To clearly distinguish definitions about the OEL problem from those about OEL solutions, we highlight the former in blue and the latter in red.

To introduce our framework, we first position it with respect to the definition of a continual reinforcement learning problem proposed in \cite{abel2023definition}. The authors formalize a class of problems where, to be optimal, an agent should never stop learning. In more details, and without strictly following their terminology, they define a learning agent as a trajectory in policy space generated by a history of interactions with the environment.
In their framework, a set of agents corresponds to the set of policies that are generated by the corresponding interaction histories and learning algorithms. A triplet formed by an environment, a performance measure, and a set of such agents defines a continual RL problem if the best performing agents in that set never converge to a single policy. That is, in a continual RL problem the best agents must continually learn. Typical problems that require these types of solutions involve non-stationary scenarios presenting ever changing features to which the agents need to continuously adapt \cite{Kauvar2023CuriousReplay, RomeroSantucci2023}.

These definitions are helpful, but a few remarks need to be made.
First, in the given definition a continual RL agent may switch between a finite set of policies forever without generating anything new. By contrast, in Section~\ref{sec:oep} we propose the idea that a key element of OEL is the never ending learning of new knowledge, building on a notion of open-ended process.

Second, the notion of continual learning given in \cite{abel2023definition} does not imply the idea of a growing set of goals that an agent may face. To account for such an idea, in Section~\ref{sec:oegcrl} we plug
our notion of open-ended process into the goal-conditioned reinforcement learning (GCRL) framework, where the open-ended process plays the role of generating goals.

\subsection{Open-ended process}
\label{sec:oep}

The key building block of our proposal is a general notion of \textit{open-ended process}. To arrive at this notion we first define a process.

\begin{defpb}
A process is an entity that generates a set of tokens through time.
\end{defpb}

There is no constraint on what a token is and no time limit, and the same token may appear several times. Note that a process may generate several tokens at the same time.

To define an \textit{open-ended process}, we introduce the idea that the process generates some novelty through time. But we would like to stress that novelty is a property relative to an observer: some observers will consider a token as new if it is different enough from all previous tokens according to their own perspective, but other observers may consider that the same token is too similar to previous tokens to be considered distinct.
The notion of observer proposed here is broad as it might be a non-transparent subjective evaluator (e.g., a human user) or alternatively an objective mechanism that incorporates some measure of novelty (e.g., an algorithm).
As a consequence, our key definition is the following:

\begin{keydef}
{\bf An observer considers a process as open-ended if, for any time $t$, there exists a time $t' > t$ at which the process generates a token that is new according to this observer's perspective.}
\end{keydef}

Differently from the definition of continual learning of \cite{abel2023definition}, this definition captures the idea that something new must happen at least from time to time. Moreover, the definition implies that an open-ended process works over an infinite time horizon.

A property of open-endedness defined this way is that any open-ended process necessarily generates an infinite number of distinct tokens.
As a simple mathematical proof, if $|{\tt{Tok}}_t|$ denotes the number of distinct tokens generated up to time $t$, by definition there exists a time $t' > t$ such that $|{\tt{Tok}}_{t'}| \geq {|\tt{Tok}}_t| + 1$. This property can be applied an arbitrary number of times, so that for any $N = 1, 2, ...$ there exists a $t'>t$ such that ${|\tt{Tok}}_{t'}| \geq |{\tt{Tok}}_t| + N$, proving that the number of distinct tokens generated throughout the process is unbounded.

The above definition is very general and can capture many examples of open-endedness. For instance, in open-ended evolution, the tokens are agent phenotypes.
In standard RL (i.e., non goal-conditioned), an OEL agent is an agent whose behavior generation process is open-ended. There might be several ways to implement this condition within our definition, for example the tokens might be reward functions, Markov decision processes, policies or trajectories. We now first give a definition of such an OEL framework based on standard RL, and then we focus on GCRL  and open-ended goal generation.

\subsection{Open-ended RL}
\label{sec:rl}

To define RL problems we borrow the formalism of the problem definitions from \cite{abel2023definition} that we introduced in Section~\ref{sec:def}.
This formalism is used here because it has a broader scope than the more frequently used Markov Decision Processes.
However, unlike in \cite{abel2023definition} where the distinction between problems and solutions is not considered, our definitions insist on this distinction.
As in \cite{abel2023definition}, we define an agent-environment ``interface'' as the $(\Ob \times \A)$ pair where $\Ob$ is the space of observations and $\A$ is the space of actions resulting from interactions of this agent with this environment.
We then define the set of possible histories of the agent-environment interactions as $\mathbf{H} = \bigcup_{t=0}^\infty (\Ob  \times \A)^t$.
A single history of such a set, $h_t \in \mathbf{H_t} \subset \mathbf{H}$, involves a sequence of length $t$ of observation-action pairs.

From these definitions, we define an RL problem as follows.

\begin{defpb}
An RL problem is a tuple $<~\Ob, \A, \T, \Rew~>$ where $\Ob$ is the set of the agent's observations, $\A$ is the set of the agent's primitive actions, $\T: \mathbf{H}_t \rightarrow \Delta(\Ob)$ is a transition function mapping the history up to time $t$ to a probability distribution of observations (where $\Delta (\Ob)$ is the probability simplex of $\Ob$), and $\Rew: \mathbf{H}_t \rightarrow \R$ is a reward function.
\end{defpb}

To build an open-ended RL problem from the RL problem as defined above, we need to consider at least a sequence of such RL problems. Such problems can be organized purely sequentially, but also hierarchically, when an agent is building complex problem representations on top of simpler representations.
In the case of a hierarchical organization, one may call upon the options framework \cite{sutton1999between}. 

Under this hierarchical perspective, the tokens generated by an open-ended process could be RL problems themselves, that is an open-ended RL problem might generate an open-ended sequence of RL subproblems.
Based on these elements, we can give the following general and potentially recursive definition of an open-ended RL problem.

\begin{keydef}
An open-ended RL problem is a sequence of RL problems in which any of the following defining elements is generated by an open-ended process: the observation space, the action space or some hierarchical options, the transition function, the reward function or RL problems.
\end{keydef}

\subsection{Open-ended GCRL}
\label{sec:oegcrl}

From these definitions, we proceed to define GCRL problems. That is, we introduce into RL problems a notion of goal and goal-conditioned reward function (see \cite{colas2022autotelic}). We characterize goals as elements of an arbitrary goal space $\G$ but we do not specify where they come from. For example, they could be generated by the environment, another agent or the agent itself.

\begin{defpb}
A GCRL problem is a tuple $<~\Ob, \A, \T, \G, \Rew~>,$ where $\Ob$ is the set of the agent's observations, $\A$ is the set of the agent's actions, $\T: \mathbf{H}_t \rightarrow \Delta(\Ob)$ is a transition function mapping the history up to time $t$ to a probability distribution of observations, $\G$ is the set of goals, and $\Rew: \mathbf{H}_t \times \G \rightarrow \R$ is the goal-conditioned reward function.
\end{defpb}

Similarly, using goals we can apply our general definition of an open-ended process to a goal generation process, and get the following definition:


\begin{keydef}
A GCRL problem is an open-ended GCRL problem if the goals come from an open-ended generation process.
\end{keydef}

As noted above, for goal-generation processes whose set of distinct goals for the observer is finite, the problem cannot be open-ended. However, in practice agents generally have a finite lifespan.
Thus a process generating a finite set of goals might be seen as giving rise to an open-ended goal generation process even if the infinite horizon condition is not met. We come back to this issue in Section~\ref{sec:eval}.

The above definitions are very general, and capture a large set of OEL problems, among which many are not satisfactory models of developmental processes. To go further towards modelling open-ended learning from a developmental perspective, we must work on distinguishing trivial cases of open-ended GCRL problems from more interesting ones.

\subsection{First-order and second-order open-ended GCRL}

Given the above definitions, there are two simplistic cases of open-ended GCRL problems.

\noindent{\bf Case 1: The goal space is discrete, but infinite.} A simple example of this case is when the goal is to count from $1$ to $N$,
where $N$ is a natural number. Generating as goals an infinite sequence of growing values of $N$ is enough for that problem to be open-ended.

\noindent {\bf Case 2: The goal space is continuous.} A simple example of this case is asking a physical agent to travel at a given speed, where the speed is a real number. Targeting higher and higher traveling speeds that converge asymptotically to the highest possible traveling speed of the agent is a problem that already verifies the open-endedness property. 

These two simplistic examples qualify as open-ended GCRL problems for naive observers who do not require a strong notion of novelty, but they do not account for the developmental perspective AI authors generally have in mind when they try to define OEL agents. We see two approaches to address this issue.

A first approach to avoid trivial OEL problems consists in putting constraints on what an observer may consider as a ``new token''  or not. Demanding observers may require new tokens to be ``substantially new'' or ``interestingly new''.

A second approach, which is not orthogonal to the previous one, consists in considering that a developmental process implies that goal tokens are sampled from more and more interesting {\em spaces}. With this second approach, we may consider two classes of problems, following a strategy inspired from the work of \cite{etcheverry2021meta}.
The first class involves problems as the usually considered ones:

\begin{defpb}
A first-order open-ended GCRL problem  is a GCRL problem where all goals $\g_i$ are sampled from the same goal space $\G$.
\end{defpb}

By contrast, to define the second class, we first need to define a variety of goal spaces $\Om_j \in \Om$ where $\Om$ is the set of all possible goal spaces. Then we can introduce the idea of goals generated from different spaces.

\begin{defpb}
A second-order open-ended GCRL problem is a GCRL problem where a first open-ended process generates an open-ended sequence of goal representation spaces $\Om_j \subset \Om$, and a second open-ended process generates goals $\g_{i,j}$ within each generated space.
\end{defpb}

Below for simplicity we consider the case of one goal space, and hence $\G$ instead of $\G_{j}$, but our definitions could be extended to the case of the generation of multiple goal spaces.

A different way to express the same thing is that in first-order GCRL problems tokens of the open-ended process are goals from a single goal space, whereas in the second-order case tokens of the open-ended processes are goal spaces and goals.

Note that with the above definition, second-order open-ended GCRL problems can still generate a trivial diversity of goal spaces, so again we have to add further conditions so that the generated goals and goal spaces ensure an interesting developmental trajectory.

\section{Combining open-ended learning with other properties}
\label{sec:combi}

We have proposed a definition for the elementary property of an OEL problem. Now it is time to ask whether and how we can combine this property with other, complementary ones. In particular, we would like to build an equivalent of the LOLA framework of \cite{romero2023perspective}, where LOLA stands for Lifelong Open-Ended Learning Autonomy.
Below we do so by first building in Section~\ref{sec:continual} on the fully-formalized notion of continual learning proposed in \cite{abel2023definition}. Then we stress the need for accounting for the lifelong learning property. Finally, we also expand in Section~\ref{sec:teachable} on the fact that the autonomous learning property is too strict a requirement, as one may consider agents that are both autotelic and teachable, i.e. capable of adopting social partner's goals.

\subsection{Goal-conditioned continual reinforcement learning}
\label{sec:continual}

Given our definitions above, we can now connect our definition of OEL to the definition of continual learning from \cite{abel2023definition}, investigating solutions to the continual and open-ended GCRL problem.
An RL agent can then be defined as follows:

\begin{soldef}
An RL agent is a function $\lambda : \mathbf{H}_t \rightarrow \Delta(\A)$ mapping a history of observations and actions up to a given time $t$, to an action selection probability distribution.
\end{soldef}

An goal-conditioned agent can then be defined as follows:

\begin{soldef}
A GCRL agent is a function $\kappa: \mathbf{H}_t \times \G \rightarrow \Delta(\A)$ mapping a history of observations and actions up to a given time $t$, and the goal selected at that time, to the action selection probability distribution.
\end{soldef}

As classically done in the GCRL literature \cite{campos2020explore}, we distinguish `behavioral goals'', which are used to condition a policy, and ``achieved goals'', which are the goals fulfilled by the behavior generated by the policy.

In \cite{abel2023definition} the authors define ``continual problems'' as problems where the optimal agents never stop learning (i.e., they continuously change their policies). A weakness of this definition is that the continual learning RL problem is defined in terms of the type of solutions (agents) to use to solve them. From that perspective, the examples given by the authors in their work all involve non-stationary environments, and thus the non-stationary property should actually be used to define their problems. Given this feature of such environments, it should follow that the optimal agents (solutions) solving them have to keep changing their policies. Furthermore, one should also consider that if the environment continues to change, but after some time it repeats the same configurations, then the best agents might actually learn all the solutions for all environment configurations, and then stop learning. This unless it is assumed that the agents forget previous solutions when the environment changes, an assumption that the authors seem to implicitly make. Only problems that continue to propose novel challenges to the agents, as the true OEL problems considered here, actually require agents that keep changing: these agents have to keep changing not because they forget previous solutions, but because they need to indefinitely acquire new knowledge. This is a signature feature of OEL agents and also of any solution to truly OEL problems.

We can define open-ended RL agents (on the side of solutions) as follows:

\begin{soldef}
An open-ended RL agent is an RL agent that solves an open-ended RL problem.
\end{soldef}

This definition is admittedly very general. We can be more specific when defining open-ended GCRL agents:

\begin{soldef}
An open-ended GCRL agent is an GCRL agent that uses goals from an open-ended goal generation process.
\end{soldef}

From this definition, it should be obvious that an open-ended GCRL agent defined as above is the solution to a continual RL problem as defined by \cite{abel2023definition}. Indeed, to be optimal, these agents should continuously learn in order to solve the continuously generated novel goals.

By contrast, if we remove the condition on OEL and if the environment is stationary, we may account for convergence to an universally efficient agent that has learned enough to become capable, after some time, to achieve any forthcoming goal in a zero-shot manner. 

Given the considerations above, we would like the agents not to (completely) forget the tasks it previously achieved when it addresses new ones. This property corresponds to lifelong learning. Thus we should also include the lifelong learning property into our framework.

\subsection{Lifelong open-ended learning}
\label{sec:lifelong}

Though the names suggest they are similar, the lifelong learning property and the continual learning property are not exactly the 
same. Indeed, the authors who use the first term generally have in mind the capability to overcome catastrophic forgetting (e.g., see \cite{parisi2019continual}) whereas the authors of the continual learning definition mentioned above do not even mention this aspect \cite{abel2023definition}.

The lifelong learning property captures the idea of a growing repertoire of skills, although it mainly focuses on avoiding to forget the previous skills when learning new ones. Thus, characterizing such a property requires being more specific on the learning processes from the side of the agent, hence the definition must be from the side of solutions.

\begin{soldef}
A lifelong open-ended RL or GCRL agent is an open-ended RL or GCRL agent which does not forget how to solve RL problems or achieve goals when it learns how to solve other RL problems or achieves other goals.
\end{soldef}


\subsection{The origin of goals: extrinsic, autotelic, and teachable open-ended learning agents}
\label{sec:teachable}

So far, we have not specified where the goals come from.
In this respect, there are three main possibilities.

First, the goals may come from the environment itself. In this case we say they are extrinsic.
This corresponds to the {\tt GoalEnv} category of environments in the standard {\tt gym} interface for RL agents.
In this case, the environment provides the agent with a state, a goal, and a reward.
Extrinsic OEL agents are thus as follows.

\begin{soldef}
An open-ended RL or GCRL agent is extrinsic if the goal generation process is external to the agent.
\end{soldef}

An important case of extrinsic OEL problem is one where the environment generates an open-ended sequence of goals that maximise the speed and quality of the learning process of the agents \cite{jiang2023general}. 
Extrinsic OEL agents, however, are not the main focus here.

The second case corresponds to agents who set and learn to pursue their own tasks or goals autonomously.
In this case we say
the goals generated by the agents are intrinsic. 
An interesting case of this is when the agents autonomously set their own goals, in which case we have autotelic agents.

\begin{soldef}
An open-ended GCRL agent is autotelic if its goal generation process is fully internal to the agent (autonomous).
\end{soldef}

To discover new goals, autotelic OEL agents generally expand the set of behavioral goals they consider by sampling from the set of goals they have already achieved \cite{campos2020explore}, which defines a curriculum over goals. More precisely, the most efficient curricula consist in sampling at the frontier of the domain of currently achieved goals \cite{seepanomwan2017intrinsically, pitis2020maximum, castanet2022stein}. This interaction should ensure that the agent achieves enough new goals to continue expanding its behavioral goal domain.

From a slightly different perspective, for an autotelic agent facing a second-order open-ended GCRL problem where it generates goal spaces from its own experience, the environment has to be rich enough to foster this open-ended goal space generation process.

A third case involves agents that autonomously generate goals but whose goal generator can be influenced by interactions with social partners. By giving demonstrations, descriptions, instructions or feedback, social partners can suggest to these agents to adopt some goals rather than others. Following \cite{sigaud2023towards}, focusing on autotelic agents, we say that such agents are teachable.

\begin{soldef}
An OEL GCRL agent is teachable if its internal goal generation process can be influenced by other agents.
\end{soldef}

It should be clear that ensuring the OEL property is more difficult for purely autonomous agents than for teachable agents, as the former must discover new goals or goal spaces without external support, whereas the latter can benefit from social interactions to do so. That is, being teachable may relax lot of the requirements on the goal discovery capabilities of agents.

Note that some agents can be teachable without autonomously generating goals, meaning that they fully depend on external agents for goal generation. Note also that autonomously generating goals is not a sufficient condition for being an OEL agent, as an autonomous agent may always sample from the same limited set of goals.

Also, it is not a sufficient condition to be a continual learner according to the definition of \cite{abel2023definition}, as the behavior may become stationary.

This line of thoughts suggests studying in the future the different properties of the types of agents developmental AI researchers have in mind, and the way they interact with each other.

\section{Evaluating open-ended learning agents}
\label{sec:eval}

The OEL framework poses two important challenges:
(1) How can we make sure that a learning agent is an OEL agent?
(2) How can we compare the performance of two OEL agents?

\subsection{Assessing the open-ended learning property}

A key difficulty with OEL agents is that, based on the definition given here, it is not easy to evaluate in practice if an agent is showing open-ended learning or not. Indeed, such an agent should address new goals forever, so we should wait forever to determine if it succeeds in doing so. Of course, this is not doable in practice. A related viewpoint is that the definition does not take into account the fact that concrete agents generally have a finite lifespan.

A first approach to address this problem was proposed by \cite{cartoni2018autonomous}, where the authors attempt to define open-ended learning problems in a concrete way.
The agent's ``life'' is divided into two phases. 
In a first, very long ``intrinsic phase'', the agent is set in a given environment and receives no learning guidance (e.g., reward functions or goals) nor any hardwired knowledge.
In a second ``extrinsic phase'', the agent is tested with a set of tasks (e.g., goals to achieve) that are ``randomly drawn'' from the same environment, in the sense that these tasks are \textit{representative of all possible tasks} that might be generated in that environment. 
The key idea is that, since in the intrinsic phase the agent does not know the tasks it will have to solve in the extrinsic phase, it will have to possibly rely on intrinsic motivations to maximise its knowledge and skill acquisition and thus have a high performance with tasks in the extrinsic phase. Importantly, this performance represents a measure of the knowledge-gain capacity of the OEL processes in the intrinsic phase.
This approach is definitely practical, but we have to be aware that it does not truly evaluate the OEL property in itself.

Another possible approach is more closely related to the definition of OEL given here. It consists in measuring the rate of novel goals that the agent discovers and/or learns to master throughout its life, and then extrapolating from its dynamics if the agent would continue to discover new goals forever if given infinite time. 
In particular, if the number of seen goals asymptotically converges to a constant, the agent is not an OEL agent. Otherwise, if the curve shows a logarithmic behavior, a linear behavior with  positive slope, or even better an increasing slope, then the agent is an OEL agent. The practical application of this approach requires the possibility of marking goals as novel in order to compute the goal-generation rate, an operation which would pose the problems previously discussed.

\subsection{Comparing open-ended learning agents}

Besides, comparing the goal-reaching capabilities of purely autotelic agents is particularly difficult since different agents may follow a completely different curriculum, thus there might be no common ground on which to compare them. However, the method proposed by \cite{cartoni2018autonomous} was designed to also address this problem. Indeed, the tasks used in the extrinsic phase to measure the agent performance should be representative and cover the whole space of problems that might be generated in the environment. Thus, the average performance of different agents on these tasks should be representative of the effectiveness of their OEL processes independently of their specific curricula which generated specific task-trajectories across the task space.

In the case of teachable agents, the problem can be mitigated as one may determine a common social interaction policy for an interacting social partner and measure to which extent various teachable agents manage to learn the goals the social partner is trying to suggest. 

\section{Discussion and Conclusion}
\label{sec:limits}

In this paper we have investigated the genealogy of the notion of open-ended learning and isolated a core property that may be general enough to account for a large number of open-ended learning phenomena. From there, we have focused on a goal-conditioned reinforcement learning perspective, and proposed a framework from which open-ended GCRL problems and agents could be defined.

In addition, we have outlined the need for combining our core property with other properties that would help developmental IA authors to capture the idea of a growing repertoire of capabilities or skills, which is so important in the OEL process of children. 

As a consequence of our choice, the main limitation of our work is that our definition of OEL problems does not imply any form of performance progress from the side of the agent. An intuition is that the agent should maximize some performance measure on the behavioral goals it has already seen so far. Thus, to go further, we should characterize a goal discovery process, explain how an agent may discover new goal spaces \cite{pong2019skew}, maximize its competence in these goal spaces \cite{Santucci2016GRAIL}, introduce representational redescription processes \cite{doncieux2018open}, abstraction capabilities \cite{konidaris2019necessity, shanahan2022abstraction}, and even creativity \cite{boden1998creativity}. 
Finding adequate performance measures for all these further capabilities is left for future work, and trying to integrate all these capabilities into a common framework may reveal some deeper limitations of the definitions we have proposed here.

\section*{Acknowledgements}
This work has received funding from the European Commission's Horizon Europe Framework Program under grant agreement $N^o$ 101070381 (PILLAR-robots project).

\end{document}